\documentclass[letterpaper]{article}
\usepackage{aaai20}
\usepackage{times}
\usepackage{helvet}
\usepackage{courier}
\usepackage[hyphens]{url}
\usepackage{graphicx}
\usepackage[]{algorithm}
\usepackage[noend]{algpseudocode}
\usepackage{graphicx}
\usepackage{amsmath, amsthm, amssymb}
\usepackage{amsfonts}

\DeclareMathOperator*{\argmin}{arg\,min}
\usepackage[utf8x]{inputenc}
\usepackage{latexsym}
\usepackage{fancyhdr}
\usepackage{lscape}
\usepackage[T1]{fontenc}
\usepackage{xspace}
\usepackage{array}

\newcommand*\samethanks[1][\value{footnote}]{\footnotemark[#1]}

\title{Knowledge Infused Learning (K-IL):\\
Towards Deep Incorporation of Knowledge in Deep Learning}
\author{Ugur Kursuncu\thanks{Equally Contributed}, 
Manas Gaur\samethanks,
Amit Sheth\\
AI Institute, University of South Carolina\\
Columbia, SC, USA\\
\{kursuncu@mailbox.sc.edu, mgaur@email.sc.edu, amit@sc.edu\}\\
}

\begin{document}
\maketitle

\begin{abstract}
Learning the underlying patterns in data goes beyond instance-based generalization to external knowledge represented in structured graphs or networks. Deep learning that primarily constitutes neural computing stream in AI has shown significant advances in probabilistically learning latent patterns using a multi-layered network of computational nodes (i.e., neurons/hidden units).  Structured knowledge that underlies symbolic computing approaches and often supports reasoning, has also seen significant growth in recent years, in the form of broad-based (e.g., DBPedia, Yago) and domain, industry or application specific knowledge graphs. A common substrate with careful integration of the two will raise opportunities to develop neuro-symbolic learning approaches for AI, where conceptual and probabilistic representations are combined. As the incorporation of external knowledge will aid in supervising the learning of features for the model, deep infusion of representational knowledge from knowledge graphs within hidden layers will further enhance the learning process. Although much work remains, we believe that knowledge graphs will play an increasing role in developing hybrid neuro-symbolic intelligent systems (bottom-up deep learning with top-down symbolic computing) as well as in building explainable AI systems for which knowledge graphs will provide scaffolding for punctuating neural computing. In this position paper, we describe our motivation for such a neuro-symbolic approach and framework that combines knowledge graph and neural networks.  
\end{abstract}

\section{Introduction}
\label{sec:intro}
Data-driven bottom-up machine/deep learning (ML) and top-down knowledge-driven approaches to creating reliable models, have shown remarkable success in specific areas, such as search, speech recognition, language translation, computer vision, and autonomous vehicles. On the other hand, they have had limited success in understanding and deciphering contextual information, such as detection of abstract concepts in online/offline human interactions. Current challenges in the translation of research methods and resources into practice often draw from a class of rarely studied problems that do not yield to contemporary bottom-up ML methods. Policymakers and practitioners assert serious usability concerns that constrain adoption, notably in high-consequence domains \cite{topol2019high}. In most cases, data-dependent ML algorithms require high computing power and large datasets, where the crucial signals may still be sparse or ambiguous, threatening precision \cite{cheng2018ai}. Moreover, the ML models that are deployed in the absence of transparency and accountability \cite{rudin2019stop} and trained on biased datasets, can lead to grave consequences, such as potential social discrimination and unfair treatment \cite{olteanu2019social}. Further, the potentially severe implications of false alarms in an ML-integrated real-world application may affect millions of people \cite{kursuncu2019modeling,kursuncu2018modeling}. 

The fundamental challenges  are common to a majority of problems in a variety of domains with real world impact. Specifically, these challenges are: (1) \emph{dependency on large datasets} required for bottom-up, data-dependent ML algorithms \cite{valiant2000robust,de2019random}, (2) \emph{bias in the dataset}, enabling the model to emph{potentially} cause social discrimination and unfair treatment, (3) \emph{multidimensionality, ambiguity} and \emph{sparsity}, as the data involves unconstrained concepts and relationships with meaning from different contextual dimensions of the content such as religion, history and politics \cite{kursuncu2019modeling,kursuncu2018modeling}. Further, the limited number of labeled instances available for training may  fail to represent the true nature of concepts and relationships in data sets, leading to ambiguous or sparse true signals (4) the lack of information traceability for model \emph{explainability}, (5) \emph{the coverage of information} specific to a domain that would be missed otherwise, (6) \emph{the complexity of model architecture in time and space}\footnote{\url{https://www.theguardian.com/commentisfree/2019/nov/16/can-planet-afford-exorbitant-power-demands-of-machine-learning}}, and (7) \emph{false alarms} in model performance. Consequently, we believe standard separate knowledge graph KG and ML methods are vulnerable to deduce or learn spurious concepts and relationships that appear deceptively good on a KG or training datasets, yet do not provide adequate results when the data set contains contextual and dynamically changing concepts and relations.

In this position paper, we describe innovations that will operationalize more abstract models built upon the characteristics of a domain to render them computationally accessible within neural network architectures.  We propose a neuro-symbolic method, \emph{knowledge-infused learning} that measures information loss in latent features learned by neural networks through KGs with conceptual and structural relationship information, for addressing the aforementioned challenges. The infusion of knowledge during the representation learning phase raises the following central research questions: (i) How do we decide whether to infuse knowledge or not, at a particular stage while learning between layers, and how to quantify knowledge to be infused? (ii) How to merge latent representations between layers with external knowledge representations, and (iii) How to propagate the knowledge through the learned latent representation? Considering the future deployment of AI in applications, the potential impact of this approach is significant. As stated in \cite{karpathy2015unreasonable}, the deeper the network, the denser the representation and better the learning. A large number of parameters and the layered nature of neural networks make them modifiable based on specific problem characteristics. However, the challenges (1, 3, 5 and 7) make neural networks vulnerable to the sudden appearance of relevant-but-sparse or ambiguous features, in often noisy big data \cite{valiant2000robust,de2019random,kursuncu}. On the other hand, KG-based approaches structure search within a  feature space defined by domain experts. To compensate for the vulnerability of the aforementioned challenges, incorporating knowledge to the learned representation in principled fashion is required. A promising approach is to base this on a measurable discrepancy between the knowledge captured in the neural network and external resources. 

Computational modeling coupled with knowledge infusion in a neural network will disambiguate important concepts defined in a KG with their different semantic meanings through its structural relations. Knowledge infusion will redefine the emphasis of sparse but essential, and irrelevant but frequently occurring terms and concepts, boosting recall without reducing precision. Further, it will provide explanatory insight into the model, robustness to noise and reduce dependency on frequency in the learning process. This neuro-symbolic learning approach will potentially transform existing methods for data analysis and building computational models. While the impact of this approach is transferable (and replicable) to a  majority of domains, the explicit implications are particularly apparent  for social science \cite{kursuncu2019modeling} and healthcare domains \cite{gaur2018let}.

\section{Related Work}
\label{sec:related}
As the incorporation of knowledge has been explored in various forms in prior research, in this section, we describe the methodologies and applications specifically related to knowledge-infused learning: Neural language models, neural attention models, knowledge based neural networks, all of which utilize external knowledge before/after the representation has been generated. 

\subsection{Neural Language Models (NLMs)}
NLMs are a category of neural networks capable of learning sequential dependencies in a sentence, and preserve such information while learning a representation. In particular, LSTM (Long Short Term Memory) networks \cite{hochreiter1997long} have emerged from the failure of RNNs (Recurrent Neural Networks) in remembering long-term information. Concerning the loss of contextual information while learning, \cite{cho2014learning} proposed a context-feed forward LSTM architecture in which context is learned by the  previous layer merged with forgetting and modulation gates of the next layer. However, if erroneous contextual information is learned in previous layers, it is difficult to correct \cite{masse2018alleviating}, which is a problem magnified by noisy data and content sparsity (e.g. Twitter, Reddit, Blogs). 

As the inclusion of structured knowledge (e.g., Knowledge Graphs) in deep learning, improves  information retrieval \cite{sheth2016semantic}, prior research has shown the significance of knowledge in the pursuit of improving NLMs, such as in commonsense reasoning \cite{liu2004commonsense}. The transformer NLMs such as BERT, \cite{devlin2018bert} (including its variants BioBert and SciBERT), are still data dependent. BERT has been utilized in hybrid frameworks such as \cite{scarlini2020sensembert} in the creation of sense embeddings using BabelNet and NASARI. \cite{liu2019k} proposed K-BERT, that enriches the representations by injecting the triples from KGs into the sentence. As this incorporation of knowledge for BERT takes place in the form of attention, we consider the K-BERT as semi-deep infusion \cite{sheth2019shades}. Similarly, ERNIE \cite{sun2019ernie} incorporated external knowledge to capture lexical, syntactic, and semantic information, enriching BERT. 

\subsection{Neural Attention Models (NAM)}
NAM \cite{rush2015neural} highlights particular features that are important for pattern recognition/classification based on a hierarchical architecture. The manipulation of attentional focus is effective in solving real world problems involving massive amounts of data \cite{halevy2009unreasonable,sun2017revisiting}. On the other hand, some applications demonstrate the limitation of attentional manipulation in a set of problems such as sentiment (mis)classification \cite{maurya2018learning} and suicide risk \cite{corbitt2016college}, where feature presence is inherently ambiguous, just as in the online radicalization problem \cite{kursuncu2019modeling}. For example, in the suicide risk prediction task, references to  suicide-related terminology appear in  social media posts of both victims as well as supportive listeners, and the existing NAMs fail to capture semantic relations between terms that  help differentiate the suicidal user from a supportive user \cite{gaur2019knowledge}. To overcome such limitations in a sentiment classification task, \cite{vo2017combination} adds sentiment scores into the feature set for enhancing the learned representation and modifies the loss function to respond to values of the sentiment score during learning. However, \cite{sheth2017knowledge,kho2019domain} have pointed out the importance of using domain-specific knowledge especially in cases where the problem is complex in nature \cite{perera2016implicit}. \cite{bian2014knowledge} has empirically demonstrated the effectiveness of combining richer semantics from domain knowledge with morphological and syntactic knowledge in the text, by modeling knowledge as an auxiliary task that  regularizes the learning of the main objective in a deep neural network.    
 
\subsection{Knowledge-based Neural Networks}
\cite{yi2018knowledge} introduced a knowledge-based, recurrent attention neural network (KB-RANN) that modifies the attentional mechanism by incorporating  domain knowledge to improve model generalization.  However, their domain-knowledge is statistically derivable from the input data itself and is analogous to merely learning an interpolation function over the existing data. \cite{dugas2009incorporating} proposed a modification in the neural network by adopting Lipschitz functions for its activation function. \cite{hu2016harnessing} proposed a combination of deep neural networks with logic rules by employing knowledge distillation procedure \cite{hinton2015distilling} of transferring the learned tacit knowledge from larger neural network, to the weights of the smaller neural network in data-limited settings. These studies for incorporating knowledge in a deep learning framework have not involved declarative knowledge structures in the form of KGs (e.g., DBpedia) \cite{chen2019embedding}. However, \cite{casteleiro2018deep} recently showed how the Cardiovascular Disease Ontology (CDO) provided context and reduced ambiguity, improving performance on a synonym detection task. \cite{shen2018knowledge} employed embeddings of entities in a KG, derived through Bi-LSTMs, to enhance the efficacy of NAMs. \cite{sarker2017explaining} presented a conceptual framework for explaining artificial neural networks’ classification behavior using  background knowledge on the semantic web. \cite{maknideep} explained a deep learning approach to learn RDFS\footnote{\url{https://www.w3.org/2001/sw/wiki/RDFS}} rules from both synthetic and real-world semantic web data. They also claim their approach improves the noise-tolerance capabilities of RDFS reasoning.

All of the frameworks in the above subsections utilized external knowledge before or after the representation has been generated by NAMs, \textit{rather than within the deep neural network} as in our approach \cite{sheth2019shades}. We propose a learning framework that infuses domain knowledge within the latent layers of neural networks for modeling.

\section{Preliminaries}
\label{sec:preliminaries}
Symbolic representation of a domain, besides its probabilistic representation, is crucial for neuro-symbolic learning. In our approach, we propose to homogenize symbolic information from KGs (see Section Knowledge Graphs) and contextual neural representations (see Section Contextual Modeling), in neural networks.  

\subsection{Knowledge Graphs}
\label{sec:KG}
A Knowledge graph (KG) is a conceptual model of a domain that stores and structures declarative knowledge in a human and machine-readable format, constituting factual ground truth and embodying a domain ontology of objects, attributes, and relations. KGs rely on symbolic propositions, employing generic conceptual relationships in taxonomies, partonomies and specific content with labeled links. Examples include DBpedia, UMLS, and ICD-10. The factual information about the domain is represented in the form of instances (or individuals) of those concepts (or classes) and relationships \cite{gruber2008ontology,sheth2012semantics}. Therefore, a domain can be described or modeled through KGs in a way that both computers and humans can understand. As KGs differentiate contextual nuances of concepts in the content, they play a key role in our framework with extensive use by several functions. 

\subsection{Contextual Modeling}
\label{sec:CM}
Capturing contextual cues in the language is crucial in our approach; hence, we utilize NLMs to generate embeddings of the content. Recent embedding algorithms have emerged  to create such representations such as Word2Vec \cite{goldberg2014word2vec}, GLoVe \cite{pennington2014glove}, FastText \cite{athiwaratkun2018probabilistic} and BERT \cite{devlin2018bert}. 
\begin{figure}
  \centering
  \includegraphics[width=0.9\linewidth]{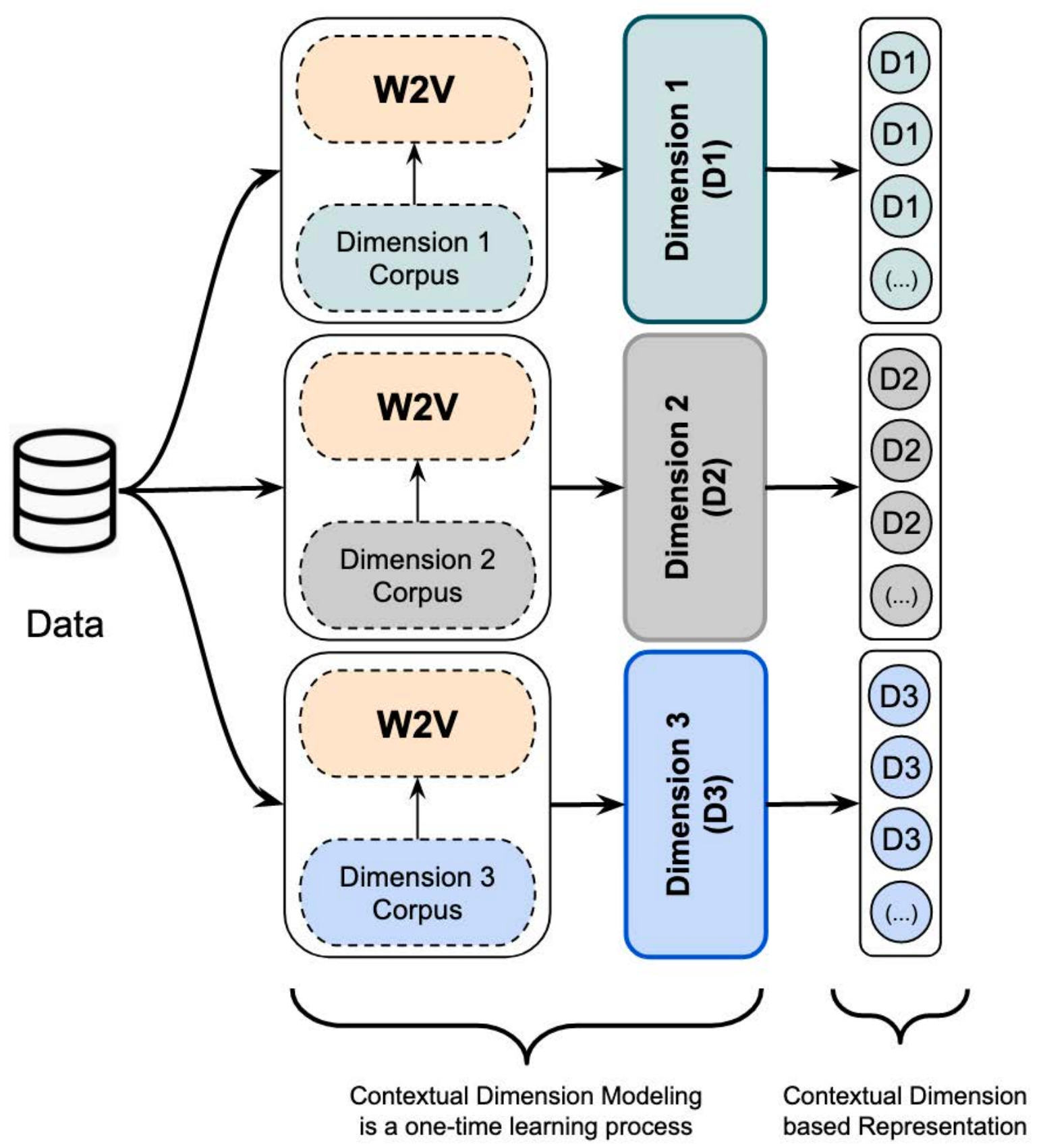}
  \caption{Contextual Dimension Modeling Diagram \cite{kursuncu2019modeling}
  Embedding algorithm above (W2V: Word2Vec) can be replaced by other algorithms such as BERT. For each dimension, a specific corpus is utilized to create the model and the generated representation of content is concatenated. Generating the three contextual dimension representations of a  social media post will emphasize the weights of such essential lexical cues.}
\label{fig:dim-mod}
\end{figure}

Modeling context-sensitive problems in different domains (e.g., healthcare, cyber social threats, online extremism and harassment), depends heavily on carefully designed features to extract meaningful information, based on characteristics of the problems and a ground truth dataset. Moreover, identifying these characteristics and differentiating the content requires different levels of granularity in the organization of features. For instance, in the problem of online Islamist extremism, the information being  shared  in  social media posts  by users  in  extremism-related social networks displays an intent that depends on the user’s type (e.g., recruiter, follower). Hence, as these user types show different characteristics \cite{kursuncu2018s}, for reliable analysis, it is critical to consider different contextual dimensions \cite{kursuncu2019modeling,kursuncu2018modeling}. Moreover, the ambiguity of diagnostic terms (e.g., jihad) also mandates representation of terms in different contexts. Hence, to better reflect these differences, creating multiple models enables us to represent the multiple contextual dimensions for a reliable analysis. Figure \ref{fig:dim-mod} details the contextual dimension modeling workflow.
 
\begin{figure*}
  \centering
  \includegraphics[width=0.9\linewidth]{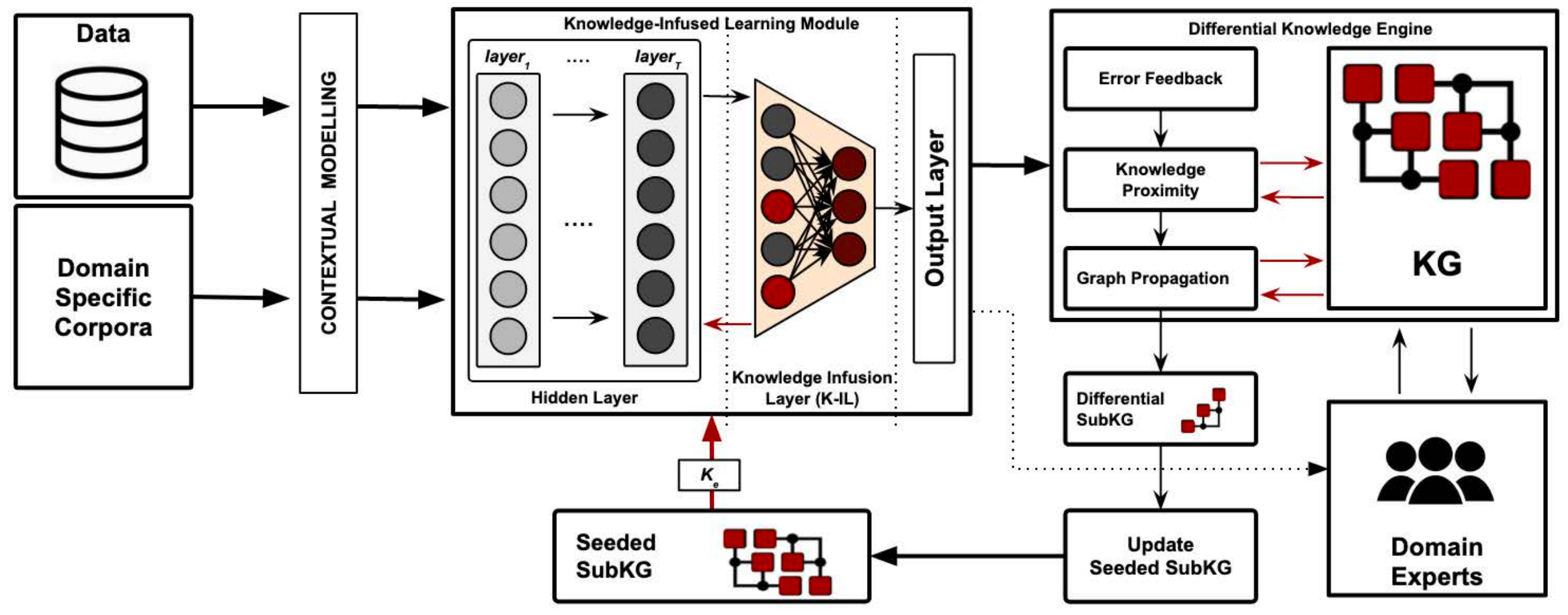}
  \caption{Overall Architecture: Contextual representations of data are generated, and domain knowledge amplifies the significance of specific important concepts that are missed in the learning model. Classification error determines the need for updating a Seeded SubKG with more relevant knowledge, resulting in a Seeded SubKG that is more refined and informative to our model.}
\label{fig:arch}
\end{figure*}

\section{A Proposed Comprehensive Approach}
\label{sec:approach}
Although  the existing research \cite{gaur2018let,bhatt2018enhancing} shows the contribution of incorporating external knowledge in ML, this incorporation mostly takes place before or after the actual learning process (e.g., feature extraction, validation); thus remaining shallow. We believe that deep knowledge infusion, within the hidden layers of neural networks, will greatly improve the performance by: (i) reducing false alarm and information loss, (ii) boosting recall without sacrificing precision, (iii) providing finer granular representation, (iv) enabling explainability \cite{islam2019infusing,kursuncu2019explainable} and (v) reducing bias. Specifically, we believe that it will become a critical and integral component of AI models that are integrated in deployed tools, e.g, in healthcare, where domain knowledge is crucial and indispensable in decision making processes. Fortunately, these domains are rich in terms of their respective machine-readable knowledge resources, such as manually curated medical KGs (e.g., UMLS \cite{mcinnes2009umls}, ICD-10 \cite{brouch2000world} and DataMed \cite{ohno2017finding}). In our prior research \cite{gaur2018let}, we utilized ML models coupled with these KGs to predict mental health disorders among 20 Mental Disorders (defined in the DSM-5) for Reddit posts. Typical approaches for such predictions employ word embeddings, such as Word2Vec, resulting in sub-optimal performance when they are used in domain-specific tasks. We have incorporated knowledge into the embeddings of Reddit posts by (i) using Zero Shot learning \cite{palatucci2009zero}, (ii) \textit{modulating} (e.g., re-weighting) their embeddings, similar to NAMs, and obtained a significant reduction in the false alarm rate, from 13\% (without knowledge) to 2.5\% (with knowledge). In another study, we have leveraged the domain knowledge in  KGs to validate model weights that explain diverse crowd behavior in the Fantasy Premier League participants (FPL) \cite{ShreyanshWI2018}. However, very little previous work has tried to integrate such functional knowledge to an existing deep  learning framework.

We propose to further develop an innovative deep knowledge-infused learning approach that will reveal patterns that are missed by traditional approaches because of sparse feature occurrence, feature ambiguity and noise. This approach will support the following integrated aims: (i) Infusion of Declarative Domain Knowledge in a Deep Learning framework, and (ii) Optimal Sub-Knowledge Graph Creation and Evolution. The overall architecture in Figure \ref{fig:arch}  guides our proposed research on these two aims. Our methods will disambiguate important concepts defined in the respective KGs with their different semantic meanings through its structural relations. Knowledge infusion will redefine the emphasis of \emph{sparse-but-essential}, and \emph{irrelevant-but-frequently-occurring} terms and concepts, boosting recall without reducing precision.

\subsection{Knowledge-Infused Learning}
\label{sec:K-IL}
Each layer in a neural network architecture produces a latent representation of the input vector ($h_t$). The infusion of knowledge during the representation learning phase raises the following central research questions: \textbf{R1:} How do we decide whether to infuse knowledge or not, at a particular stage while learning between layers, and how to quantify knowledge to be infused? \textbf{R2:} How to merge latent representations between layers with external knowledge representations, and \textbf{R3:} How to propagate the knowledge through the learned latent representation? We propose to define two functions to address these two questions: \textit{Knowledge-Aware Loss Function (K-LF)} and \textit{Knowledge Modulation Function (K-MF)}, respectively.

Configurations of neural networks can be designed in various ways depending on the problem. As our aim is to infuse knowledge within the neural network, such an operation can take place (i) before the output layer (e.g., SoftMax), (ii) between hidden layers (e.g., reinforcing the gates of an NLM layer, modulating the hidden states of NLM layers, Knowledge-driven NLM dropout and recurrent dropout between layers). To illustrate (i), we describe our initial approach to neural language models that infuses knowledge before the output layer, which we believe will shed the light towards a reliable and robust solution with more research and rigorous experimentations.

\subsubsection{Seeded Sub-Knowledge Graph}
The Seeded Sub-Knowledge Graph, is a subset of  KGs, which  participate broadly in our technical approach. Generic KGs (e.g., DBpedia \cite{bizer2009dbpedia}, YAGO2 \cite{hoffart2013yago2}, Freebase \cite{bollacker2008freebase}) may contain over a million entities and close to a billion relationships. Using the entire graph of linked data on the web can cause; (1) unnecessary computation and (2) noise due to irrelevant knowledge, and has sometimes failed to benefit intelligent application \cite{roy2017learning}. However,  real-world problems are domain-specific and require only a relevant (sub) portion of the full graph. Creation of a Seeded Sub-KG \cite{lalithsena2018domain} based on a ground truth dataset is needed, to represent a particular domain using information-theoretic approaches (e.g., KL divergence) and probabilistic soft logic \cite{kimmig2012short}. Further, a sub-graph discovery approach \cite{cameron2015context,lalithsena2018domain} can also be used utilizing probabilistic graphical models (e.g., deep belief networks, conditional random fields). In our approach, the Seeded SubKG will be updated with more knowledge based on difference between the learned representation and relevant knowledge representation from the KG (see Section Differential Knowledge Engine).

\subsubsection{$K_e$: Knowledge Embedding Creation}
Representation of knowledge in the Seeded SubKG will be generated as embedding vectors. Specific contextual dimension models and/or more generic models can be utilized to create an embedding of each concept and their relations in the Seeded SubKG. Unlike traditional approaches that compute the representation of each concept in the KGs by simply taking an average of embedding vectors of concepts, we leverage the existing structural information of the graph. This procedure is formally defined:  

\begin{equation}
\label{eq:1}
K_e = \sum_{ij} [C_i, C_j] \bigotimes D_{ij}
\end{equation}

where $K_e$ is the representation of the concepts enriched by the relationships in the Seeded-KG, ($C_i$, $C_j$) is the relevant pair of concepts in the Seeded-KG, $\mathbf{D_{ij}}$ is the distance measure (e.g., Least Common Subsumer \cite{baader2007computing}) between the two concepts $C_i$ and $C_j$. Novel methods will further be examined building upon this initial approach above as well as existing tools that include TRANS-E \cite{bordes2013translating}, TRANS-H \cite{wang2014knowledge}, and HOLE \cite{nickel2016holographic} for the creation of embeddings from KGs.

\subsubsection{Knowledge Infusion Layer}
In a many-to-one NLM \cite{shivakumar2018learning} network with $\mathbf{T}$ hidden layers, the $\mathbf{T^{th}}$ layer contains the learned representation before the output layer. The output layer (e.g., SoftMax) of the NLM model estimates the error to be back-propagated. As the techniques for knowledge infusion between hidden layers or just before the output layers will be explored, in this subsection, we explain the Knowledge Infusion Layer (\mbox{K-IL}) which takes place just before the output layer.

\begin{algorithm}[H]
\footnotesize
\begin{algorithmic}[1]
\Procedure{KnowledgeInfusion}{}
 \State $Data : NLM_{type}, \#Epochs, \#Iter, K_e$
 \State $Output:\overrightarrow{M_T}$
 \For{ne=1 to \#Epochs}
 \State $\overrightarrow{h_T}$, $\overrightarrow{h_{T-1}}$ $\leftarrow$ TrainingNLM($NLM_{Type}$,$\#Iter$)
 \While{($\textbf{D}_{KL}(\overrightarrow{h_{T-1}}||\overrightarrow{K_e})-
 \textbf{D}_{KL}(\overrightarrow{h_T}||\overrightarrow{K_e}) > \epsilon)$}
 \State ${h_T}$ $\leftarrow$ $\sigma(W_{hk}*(\overrightarrow{{h_T}}\oplus \overrightarrow{K_e})+b_{hk})$
 \State $W^{hk}$ $\leftarrow$ $W^{hk}$ - $\eta_{k}\nabla (\mbox{K-LF})$
 \EndWhile
 \State $\overrightarrow{M_T}$ $\leftarrow$ $\overrightarrow{h_T}$ $\odot$ $W^{hk}$
 \EndFor
 \State $\textbf{return:}$ $\overrightarrow{M_T}$
 \EndProcedure
\end{algorithmic}
\caption{Routine for Infusion of Knowledge in NLMs}
\label{alg:1}
\end{algorithm}

Algorithm \ref{alg:1} takes the type of neural language model, number of epochs, iterations and the seeded knowledge graph embedding $\mathbf{K_e}$ as input, and returns a knowledge infused representation of the hidden state $\mathbf{M_T}$. In line 4, the infusion of knowledge takes place after each epoch without obstructing the learning of the vanilla NLM model and is explained in lines 5-10. Within the knowledge infusion process (lines 7-9), we optimize the loss function in equation \ref{eq:2} with convergence condition defined as the reduction in the  difference between the $\mathbf{D_{KL}}$ of $\mathbf{h_T}$ and $\mathbf{h_{T-1}}$ in the presence of $K_e$. Considering the vanilla structure of a NLM \cite{greff2017lstm}, $\mathbf{M_T}$ is utilized by the fully connected layer for classification.

\begin{figure}
  \centering
  \includegraphics[width=\linewidth]{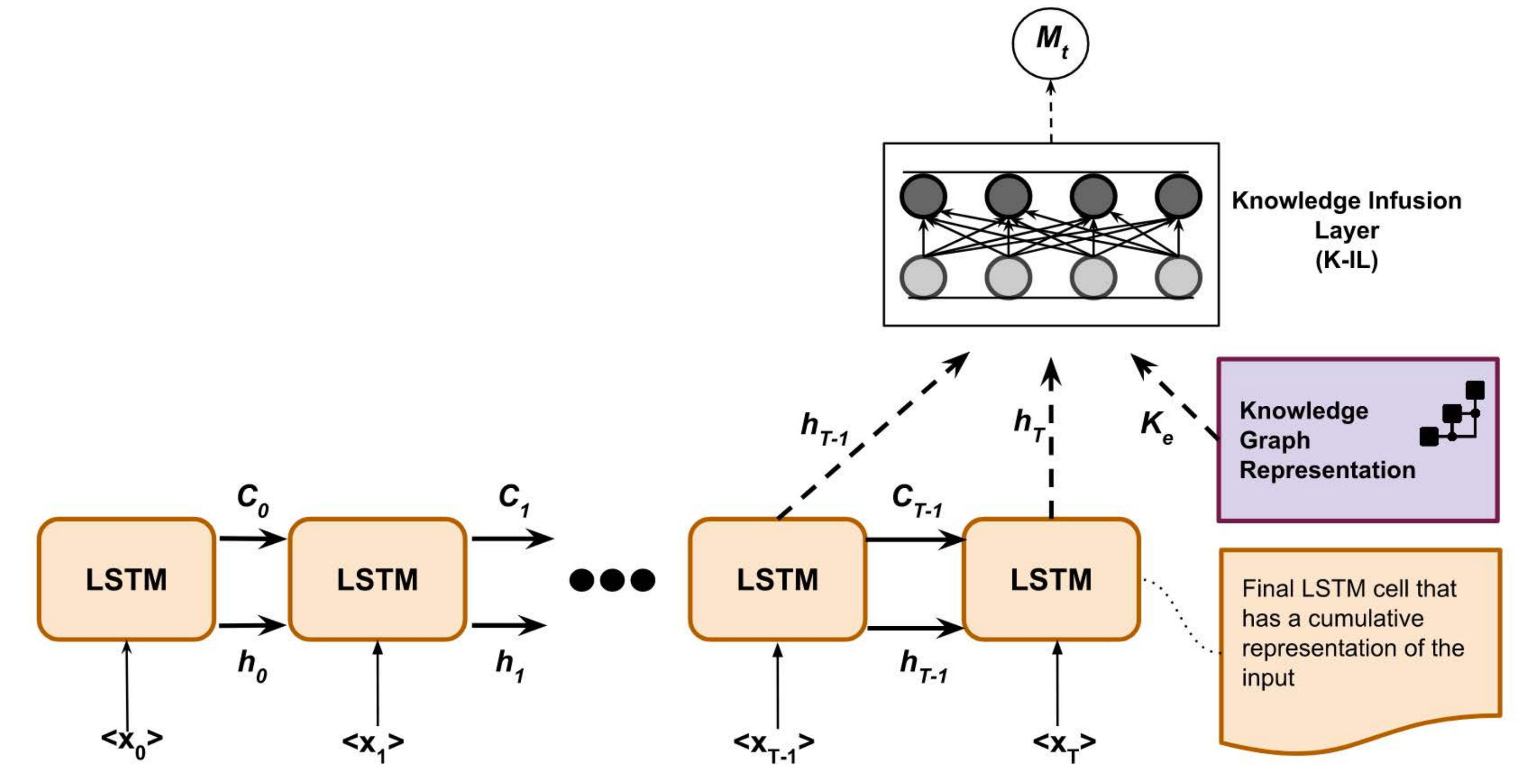}
\caption{Inner Mechanism of the Knowledge Infusion Layer}
\label{fig:kil-inner}
\end{figure}

To illustrate an initial approach in Figure \ref{fig:kil-inner}, we use LSTMs as NLMs in our neural network. \mbox{K-IL} functions  add an additional layer before the output layer of our proposed neural network architecture. This layer takes the latent vector ($\mathbf{h_{T-1}}$) of the penultimate layer, the latent vector of the last hidden layer ($\mathbf{h_T}$) and the knowledge embedding ($K_e$), as input.  

In this layer, we define two particular functions that will be critical for merging the latent vectors from the hidden layers and the knowledge embedding vector from the KG. Note that the dimensions of these vectors are the same because they are created from the same models (e.g., contextual models), which makes the merge operation of those vectors possible and valid.

\subsubsection{\mbox{K-LF}: Knowledge-Aware Loss Function}
In neural networks, hidden layers may de-emphasize important patterns due to the sparsity of certain features during learning, which causes information loss. In some cases, such patterns may not even appear in the data. However, such relations or patterns may be defined in KGs with even more relevant knowledge. We call this information gap between the learned representation of the data and knowledge representation as \textbf{differential knowledge}. Information loss in a learning process is relative to the distribution that suffered the loss. Hence, we propose a measure to determine the \textit{differential knowledge} and guide the degree of knowledge infusion in learning. As our initial approach to this measure, we developed a two-state regularized loss function by utilizing Kullback Leibler (KL) divergence. Our choice of KL divergence measure is largely influenced by the Markov assumptions made in language modeling and have been highlighted in \cite{longworth2010kernel}. The $\mbox{K-LF}$ measure estimates the divergence between the hidden representations ($\mathbf{h_{T-1},h_T}$) and knowledge representation ($K_e$), to determine the differential knowledge to be infused.

Formally we define it as: \\
$\argmin(\vec{h_{T-1}}, \vec{h_T}, \vec{K_e}) \equiv \mbox{K-LF}$, where $\mathbf{h_{T-1}}$ is an input for convergence constraint.

\begin{equation}
\label{eq:2}
\begin{split}
 \mathbf{\mbox{K-LF}} = min \ \mathbf{D}_{KL}(\vec{h_T}||\vec{K_e}); \\
 s.t. \mathbf{D}_{KL}(\vec{h_T}||\vec{K_e}) < \mathbf{D}_{KL}(\vec{h_{T-1}}||\vec{K_e})
\end{split}
\end{equation}
\\

We minimize the \textit{relative entropy} for information loss to maximize the information gain from the knowledge representation (e.g., $K_e$). We will compute differential knowledge ($\nabla\mathbf{\mbox{K-LF}}$) through such optimization approach; thus, the computed differential knowledge will also determine the degree of knowledge to be infused. $\nabla\mathbf{\mbox{K-LF}}$  will be computed in the form of embedding vectors, and the dimensions from $K_e$ will be preserved.

\subsubsection{\mbox{K-MF}: Knowledge Modulation Function}
We need to merge the differential knowledge representation with the partially learned representation. However, this operation cannot be done arbitrarily as the vector spaces of both representations are different both in dimension and distribution if not same \cite{dumanvcic2017demystifying}. We explain an  initial approach for the \mbox{K-MF} to modulate the learned weight matrix of the neural network with the hidden vector through an appropriate operation (e.g., Hadamard pointwise multiplication). This operation at the $\mathbf{T^{th}}$ layer can be formulated as:

Equation for $W^{hk} = W^{hk} - \eta_k * \nabla\mathbf{\mbox{K-LF}}$, \ where $W^{hk}$ is the learned weight matrix infusing knowledge, $\eta_k$ is learning momentum \cite{sutskever2013importance}, $\nabla\mathbf{\mbox{K-LF}}$ is differential knowledge. The weight matrix ($W^{hk}$) is computed through the learning epochs utilizing the differential knowledge embedding ($\nabla\mathbf{\mbox{K-LF}}$). Then we merge $W^{hk}$ with the hidden vector $\mathbf{h_T}$  through the K-MF. Considering that we use Hadamard pointwise multiplication as our initial approach, we formally define the output $\mathbf{M_T}$ of \mbox{K-MF} as:

This operation at the $\mathbf{T^{th}}$ layer can be formulated as:
\begin{equation}
    \vec{M_T} = \vec{h_T} \odot W^{hk}
\end{equation}

where $\mathbf{M_T}$ is Knowledge-Modulated representation, $\mathbf{h_T}$ is the hidden vector and $W^{hk}$ is the learned weight matrix infusing knowledge. Further investigations of techniques for \mbox{K-MF} constitutes a central  research topic  for the research community.

\subsection{Differential Knowledge Engine}
\label{sec:diffKE}
In deep neural networks, each epoch generates an error that is back-propagated until the model reaches a saddle point in the local minima, and the error is reduced in each epoch. The error indicates the difference between probabilities of actual and predicted labels, and this difference can be used to enrich the Seeded SubKG in our proposed knowledge-infused learning (\mbox{K-IL}) framework.

In this subsection, we discuss the sub-knowledge graph operations that are based on the difference between the learned representation of our knowledge-infused model ($\mathbf{M_T}$), and the representation of the relevant sub-knowledge graph from the KG, which we name the  differential sub-knowledge graph. We define a \textit{Knowledge Proximity function} to generate the \textit{Differential Sub-knowledge Graph}, and \textit{Update Seeded SubKG} to insert the differential sub-knowledge graph into the Seeded SubKG.

\subsubsection{Knowledge Proximity}
Upon the arrival of the learned representation from the knowledge-infused learning model, we query the KG for retrieving related information to the respective data point. In this particular step, it is important to find the optimal proximity between the concept and its related concepts. For example, from the “South Carolina'' concept, we may traverse the surrounding concepts with a varying  number of hops (empirically decided). Finding the optimal number of hops towards each direction from the concept in question is still an open research question. As we find optimal proximity of a particular concept in the KG, we propagate KG based on the proximity starting from the concept in question.

\subsubsection{Differential SubKG}
Once we obtain the SubKG from the graph propagation, we create a differential SubKG that will reflect the difference in knowledge from the Seeded SubKG. For this procedure, research is needed to formulate the problem using variational autoencoders to extract a \textit{differential subKG}($\mathbf{D_{kg}}$) and, we believe it will provide missing information in the Seeded-KG.

\subsubsection{Update function}
The differential subKG generated as a result of minimizing knowledge proximity is considered as an input factual graph to the update procedure. As a result, the procedure dynamically evolves the Seeded subKG with missing information from differential subKG. We propose to utilize \textit{Lyapunov stability theorem } \cite{liu2014attribute} and \textit{Zero Shot learning} to update the Seeded-KG using $D_{kg}$.  $D_{kg}$ and Seeded-KG represent two knowledge structures requiring a process of transferring the knowledge from one structure to another \cite{hamaguchi2017knowledge}. We define this process as generation of semantic mapping weights that encodes and decodes the two semantic spaces, utilizing the Lyapunov stability constraint and Sylvester optimization approach:
Given two semantic spaces belonging to a domain D (e.g., online extremism, mental health), we tend to attain an equilibrium position defined as:

\begin{equation}
\label{eq:4}
    ||S_{kg} - W*D_{kg}||^2_F = \alpha*||W*S_{kg} - D_{kg}||^2_F
\end{equation}

$\vert\vert\,.\,\vert\vert_F$ represents Frobenius norm and $\alpha$ is a proportionality constant belong to $\mathbb{R}$. Equation \ref{eq:4} reflects Lyapunov stability theorem and to achieve such a stable state we define our optimization function as follows:

\begin{equation}
\label{eq:5}
\begin{split}
    L = min ( ||S_{kg} - WD_{kg}||^2_F  - \alpha*||WS_{kg} - D_{kg}||^2_F),\\
    \alpha > 0,  W \in \mathbb{R} X \mathbb{R}   
\end{split}
\end{equation}

Equation \ref{eq:5} is solvable using Sylvester optimization and its derivation is defined in a recent study  \cite{gaur2018let}.


\section{Applications for \mbox{K-IL}}
\label{sec:applications}
Artificial intelligence models will be widely deployed in real world decision making processes in the foreseeable future, once  the challenges described in Section 1, are overcome. As we argue that the incorporation of external structured knowledge will address these challenges, it will benefit various application domains such as social and health sciences, automating processes that require knowledge and intelligence. Specifically, it will have a potentially significant impact on predictive analysis of online communications such as misinformation and extremism, conversational modeling, and disease prediction. 

As predicting online extremism is challenging and false alarms create serious implications potentially affecting millions of individuals, \cite{kursuncu2019modeling} showcased that the (shallow) infusion of external domain-specific knowledge improves precision, reducing potential social discrimination. Further, in prediction of mental health diseases defined in DSM-5, \cite{gaur2018let} shallow knowledge infusion reduces  false alarms by 30\%. On the other hand, conversational models pose an important application area as \cite{liu2019knowledge} proposed a conversation framework where the fusion of KGs and text mutually reinforce each other to generate knowledge-aware responses, improving the model in generalizability and explainability. In another study, \cite{young2018augmenting} integrated commonsense knowledge into the conversational models selecting the most appropriate response. While machine learning finds many application areas in medicine for disease prediction, large data is not always available. In this  case knowledge-infused learning generates more representative features thereby avoiding overfitting. A study \cite{tan2019expert} on early diagnosis of lung cancer using computed tomography  images, infused knowledge in the form of expert-curated features into the learning process through CNN. Despite the small data set, the enriched feature space in their knowledge-infused learning process improved sensitivity and specificity of the model.  

In contrast to the applications above, we believe that the deep infusion of external knowledge within latent layers will enhance the coverage of the information being learned by the model based on KGs. Hence, this will provide better generalizability, reduction in bias and false alarms, disambiguation, less reliance on large data, explainability, reliability and robustness, to the real world applications in critical aforementioned domains with significant impact.

\section{Conclusion}
\label{sec:conclusion}
Combining deep learning and knowledge graphs in a hybrid neural-symbolic learning framework will further enhance performance and accelerate the convergence of the learning processes. Specifically, the impact of this improvement in very sensitive domains such as health and social science, will be significant with respect to their implications for  real-world deployment. Adoption of the tools that automate tasks that require knowledge and intelligence, and are traditionally done by humans, will improve with the help of this framework that marries deep learning and knowledge graph techniques. Specifically, we envision that the infusion of knowledge as described in this framework will capture information for the corresponding domain in finer granularity of abstraction. We believe that this approach will provide reliable solutions to the problems faced in deep learning, as described in Sections 1 and 5. Hence, in real world applications, resolving these issues with both knowledge graphs and deep learning in a hybrid neuro-symbolic framework will greatly contribute to fulfilling AI’s promise.

\section*{Acknowledgement}
We acknowledge partial support from the National Science Foundation (NSF) award CNS-1513721: “Context-Aware Harassment Detection on Social Media". Any opinions, conclusions or recommendations expressed in this material are those of the authors and do not necessarily reflect the views of the NSF.

\bibliography{references}
\bibliographystyle{aaai}
\end{document}